\documentclass[pdflatex,sn-mathphys]{sn-jnl}
\jyear{2023}
\usepackage[sectionbib]{bibunits}%
\defaultbibliographystyle{unsrt} 
\defaultbibliography{bibliography}
\begin{document}

\title[Global AI Flood Forecasting]{AI Increases Global Access to Reliable Flood Forecasts}

\author[1]{\fnm{Grey} \sur{Nearing}}\email{nearing@google.com}
\author[1]{\fnm{Deborah} \sur{Cohen}}
\author[1]{\fnm{Vusumuzi} \sur{Dube}}
\author[1]{\fnm{Martin} \sur{Gauch}}
\author[1]{\fnm{Oren} \sur{Gilon}}
\author[2]{\fnm{Shaun} \sur{Harrigan}}
\author[1]{\fnm{Avinatan} \sur{Hassidim}}
\author[3]{\fnm{Daniel} \sur{Klotz}}
\author[1]{\fnm{Frederik} \sur{Kratzert}}
\author[1]{\fnm{Asher} \sur{Metzger}}
\author[4]{\fnm{Sella} \sur{Nevo}}
\author[2]{\fnm{Florian} \sur{Pappenberger}}
\author[2]{\fnm{Christel} \sur{Prudhomme}}
\author[1]{\fnm{Guy} \sur{Shalev}}
\author[1]{\fnm{Shlomo} \sur{Shenzis}}
\author[1]{\fnm{Tadele} \sur{Tekalign}}
\author[1]{\fnm{Dana} \sur{Weitzner}}
\author[1]{\fnm{Yossi} \sur{Matias}}

\affil[1]{\orgdiv{Google}}
\affil[2]{\orgdiv{European Centre for Medium-Range Weather Forecasts}}
\affil[3]{\orgdiv{Institute for Machine Learning, Johannes Kepler University}}
\affil[4]{\orgdiv{RAND Corporation, work was done while at Google}}

\abstract{
Floods are one of the most common  natural disasters, with a disproportionate impact in developing countries that often lack dense streamflow gauge networks. Accurate and timely warnings are critical for mitigating flood risks, but hydrological simulation models typically must be calibrated to long data records in each watershed. Using AI, we achieve reliability in predicting extreme riverine events in ungauged watersheds at up to a 5-day lead time that is similar to or better than the reliability of nowcasts (0-day lead time) from a current state of the art global modeling system (the Copernicus Emergency Management Service Global Flood Awareness System). Additionally, we achieve accuracies over 5-year return period events that are similar to or better than current accuracies over 1-year return period events. This means that AI can provide flood warnings earlier and over larger and more impactful events in ungauged basins. The model developed in this paper was incorporated into an operational early warning system that produces publicly available (free and open) forecasts in real time in over 80 countries. This work highlights a need for increasing the availability of hydrological data to continue to improve global access to reliable flood warnings.
}


\maketitle

\section{Flood forecasting is limited by data} \label{sec:introduction}
Floods are the most common type of natural disaster \cite{unisdr2015human}, and the rate of flood-related disasters has more than doubled since the year 2000 \cite{wmo20212021}. This increase in flood-related disasters is driven by an accelerating hydrological cycle caused by anthropogenic climate change \cite{milly2002increasing, tabari2020climate}. Early warning systems are an effective way to mitigate flood risks, reducing flood-related fatalities by up to 43\% \cite{world2014global,wmo2013global} and economic costs by 35-50\% \cite{pilon2002guidelines,rogers2011costs}. Populations in low- and middle-income countries make up almost 90\% of the 1.8 billion people vulnerable to flood risks \cite{rentschler2022flood}. The World Bank estimated that upgrading flood early warning systems in developing countries to the standards of developed countries would save an average of 23,000 lives per year \cite{hallegatte2012cost}. 

In this paper, we evaluate the extent to which artificial intelligence (AI) trained on open, public data sets can be used to improve global access to forecasts of extreme events in golbal rivers. Based on the model and experiments described in this paper, we developed an operational system that produces short term (7-day) flood forecasts in over 80 countries. These forecasts are available in real time without barriers to access like monetary charge or website registration \textbf{(\url{https://g.co/floodhub})}.

A major challenge for riverine forecasting is that hydrological prediction models must be calibrated to individual watersheds using long data records \cite{razavi2013anefficient,LI2010378}. Watersheds that lack stream gauges to supply data for calibration are called \textit{ungauged basins}, and the problem of `Prediction in Ungauged Basins' (PUB) was the decadal problem of the International Association of Hydrological Sciences (IAHS) from 2003--2012 \cite{sivapalan2003iahs}. At the end of the PUB decade, the IAHS reported that little progress had been made against the problem, stating that \textit{``much of the success so far has been in gauged rather than in ungauged basins, which has negative effects in particular for developing countries"} \cite{hrachowitz2013decade}.

\begin{figure}[h]
    \centering
    \includegraphics[width=89mm]{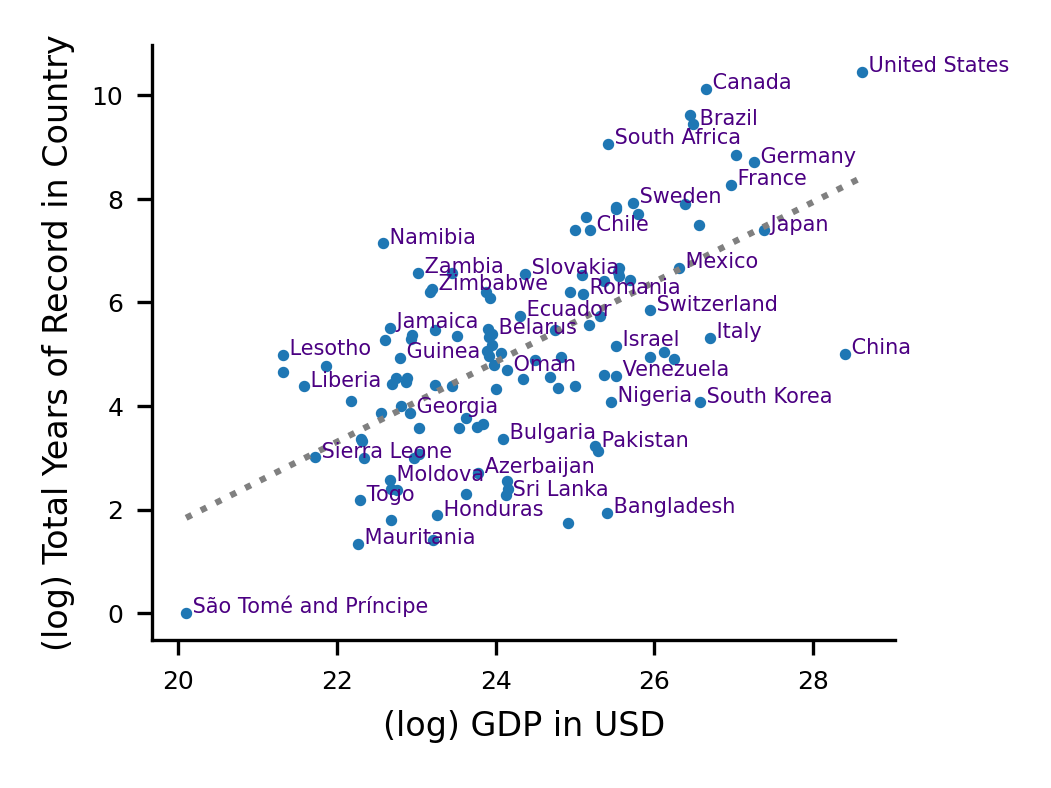}
    \caption{There is a log-log correlation (r=0.611; N=117) between national Gross Domestic Product (GDP) and the total number of years worth of daily streamflow data available in a country from the Global Runoff Data Center. GDPs are sourced from The World Bank \cite{twb2023gdp}.}
    \label{fig:gauge-gdp-correlation}
\end{figure}

Only a few percent of the world's watersheds are gauged, and stream gauges are not distributed uniformly across the world. There is a strong correlation between national GDP and the total publicly available streamflow observation data record in a given country (Figure \ref{fig:gauge-gdp-correlation} in Extended Data shows this log-log correlation), which means that high quality forecasts are especially challenging in areas that are most vulnerable to the human impacts of flooding.

In previous work \cite{kratzert2019toward}, we showed that machine learning (ML) can be used to develop hydrological simulation models that are transferable to ungauged basins. Here we develop that into a global scale forecasting system with the goal of understanding scalability and reliability. This paper addresses the following question: \textbf{Given the publicly available global streamflow data record, is it possible to provide accurate river forecasts across large scales, especially of extreme events, and how does this compare to current state of the art?} 

The current state of the art for real time, global scale hydrological prediction is the Global Flood Awareness System (GloFAS) \cite{alfieri2013glofas, harrigan2023daily}. GloFAS is the global flood forecasting system of Copernicus Emergency Management Service (CEMS), delivered under the responsibility of the European Commission’s Joint Research Centre (JRC) and operated by the European Centre for Medium-Range Weather Forecasts (ECMWF) in its role of CEMS hydrological Forecast Computation Centre. We use GloFAS version 4, which is the current operational version that went live in July, 2023. Other forecasting systems exist for different parts of the world \citep[e.g.,][]{arheimer2020global, souffront2019hydrologic, sheffield2014drought}, and many countries have national agencies responsible for producing early warnings. Given the severity of impacts that floods have on communities around the world, we consider it critical that forecasting agencies evaluate and benchmark their predictions, warnings, and approaches, and an important first step toward this goal is archiving historical forecasts.

\section{AI improves reliability and lead times of forecasts of extreme events in rivers globally} \label{sec:results}
The AI model developed for this study uses Long Short-Term Memory (LSTM) networks \citep{hochreiter1997long} to predict daily streamflow through a 7-day forecast horizon. The model is described in detail in Methods, and a version of the model suitable for research is implemented in the open source NeuralHydrology repository \cite{kratzert2022neuralhydrology}. Input, target, and evaluation data are described in Methods. 

This AI forecast model was trained and tested out-of-sample using random k-fold cross validation across 5,680 streamflow gauges. Other types of cross validation experiments are reported in Extended Data (i.e., by withholding all gauges in terminal watersheds, entire climate zones, or entire continents). Additionally, all metrics reported for the AI model were calculated with streamflow gauge data from time periods not present in training (in addition to stream gauges that were not present in training), meaning that cross-validation splits were out-of-sample across time and location. In contrast, metrics for GloFAS were calculated over a combination of gauged and ungauged locations, and over a combination of calibration and validation time periods. This means that the comparison favors the GloFAS benchmark. This is necessary because calibrating GloFAS is computationally expensive to the extent that it is not feasible to re-calibrate over cross-validation splits.

Our objective is to understand the reliability of forecasts of extreme events, so we report precision, recall, and F1 scores (F1 scores are the harmonic mean of precision and recall) over different return period events. Other standard hydrological metrics are reported in Extended Data. Statistical tests are described in Methods.

\begin{figure}[ht]
    \centering
    \includegraphics[width=120mm]{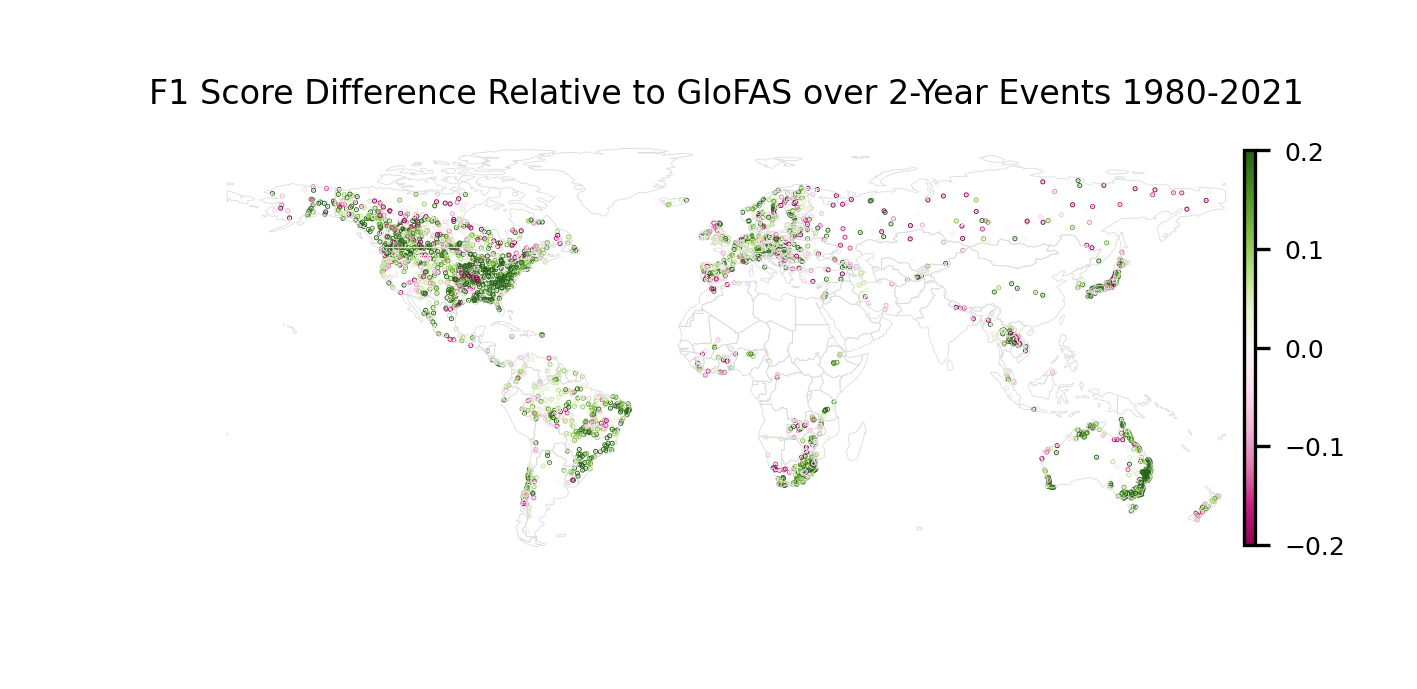}
    \caption{Differences between nowcast (0-day lead time) F1 scores for 2-year return period events between our AI model and GloFAS over the time period 1984--2021. The AI model improves over GloFAS in 70\% of gauges ($N=3,673$).}
    \label{fig:global-f1-scores-map}
\end{figure}

Figure \ref{fig:global-f1-scores-map} shows the global distribution of F1 score differences for 2-year return period events at a 0-day lead time over the time period 1984--2021 (N=3,360). Lead time is expressed as the number of days from the time of prediction, such that a 0-day lead time means that streamflow predictions are for the current day (nowcasts). The AI model improved over (was at least equivalent to) GloFAS version 4 in 64\% (65\%), 70\% (73\%), 60\% (73\%), and 49\% (76\%) of gauges for return period events of 1-year ($N=3,638$, $p=6e-87$, $d=0.22$), 2-years ($N=3,673$, $p<3e-181$, $d=0.41$), 5-years ($N=3,360$, $p=8e-130$; $d=0.42$), and 10-years ($N=2,920$, $p<1e-66$, $d=0.33$). 

\subsection{Return periods} \label{sec:results:return-periods}
More extreme hydrological events (i.e., events with larger return periods) are both more important and (when using classical hydrology models) typically more difficult to predict. A common concern \cite[e.g.,][]{sellars2018grand, todini2007hydrological, herath2021hydrologically, reichstein2019deep} about using AI or other types of data-driven approaches is that reliability might degrade over events that are rare in the training data. There is prior evidence that this concern might not be valid for streamflow modeling \cite{hess-26-3377-2022}. 

\begin{figure}[ht]
    \centering
    \includegraphics[width=89mm]{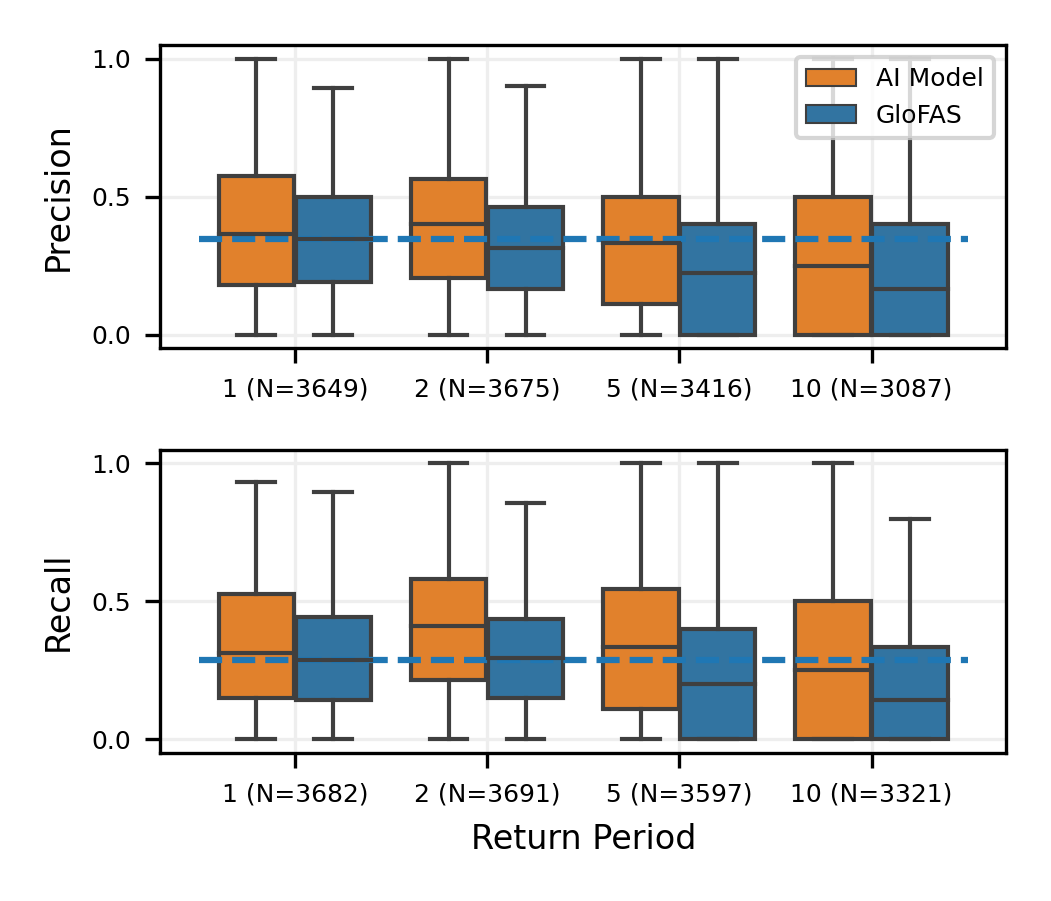}
    \caption{Distributions over (top) precision, (bottom) recall for 0-day lead time at all gauges as a function of return period. The AI model is more reliable, on average, over all return periods. Over 5-year return period events, the AI model has precision that is not statistically different than and recall that is better than GloFAS over 1-year return period events. Statistical tests are reported in the main text. Boxes show distribution quartiles and whiskers show the full range excluding outliers. The blue dashed line is the median score for GloFAS over 1-year events, and is plotted as a reference.}
    \label{fig:global-reliability-with-return-period}
\end{figure}

Figure \ref{fig:global-reliability-with-return-period} shows distributions over precision and recall for different return period events. The AI model has higher precision and recall scores for all return periods ($N>3,000$, $p<1e-5$), with effect sizes ranging from $d=0.15$ (1-year precision scores) to $d=0.46$ (2-year recall scores). Differences between precision scores from the AI model over 5-year return period events and from GloFAS over 1-year return period events are not significant at $\alpha=1\%$ ($N=3,465$, $p=0.02$, $d=-0.01$), and recall scores from the AI model for 5-year events are better than GloFAS recall scores for 1-year events ($N=3,586$, $p=1e-18$, $d=0.20$).

\subsection{Forecast lead time} \label{sec:results:lead-time}
\begin{figure}[ht]
    \centering
    \includegraphics[width=120mm]{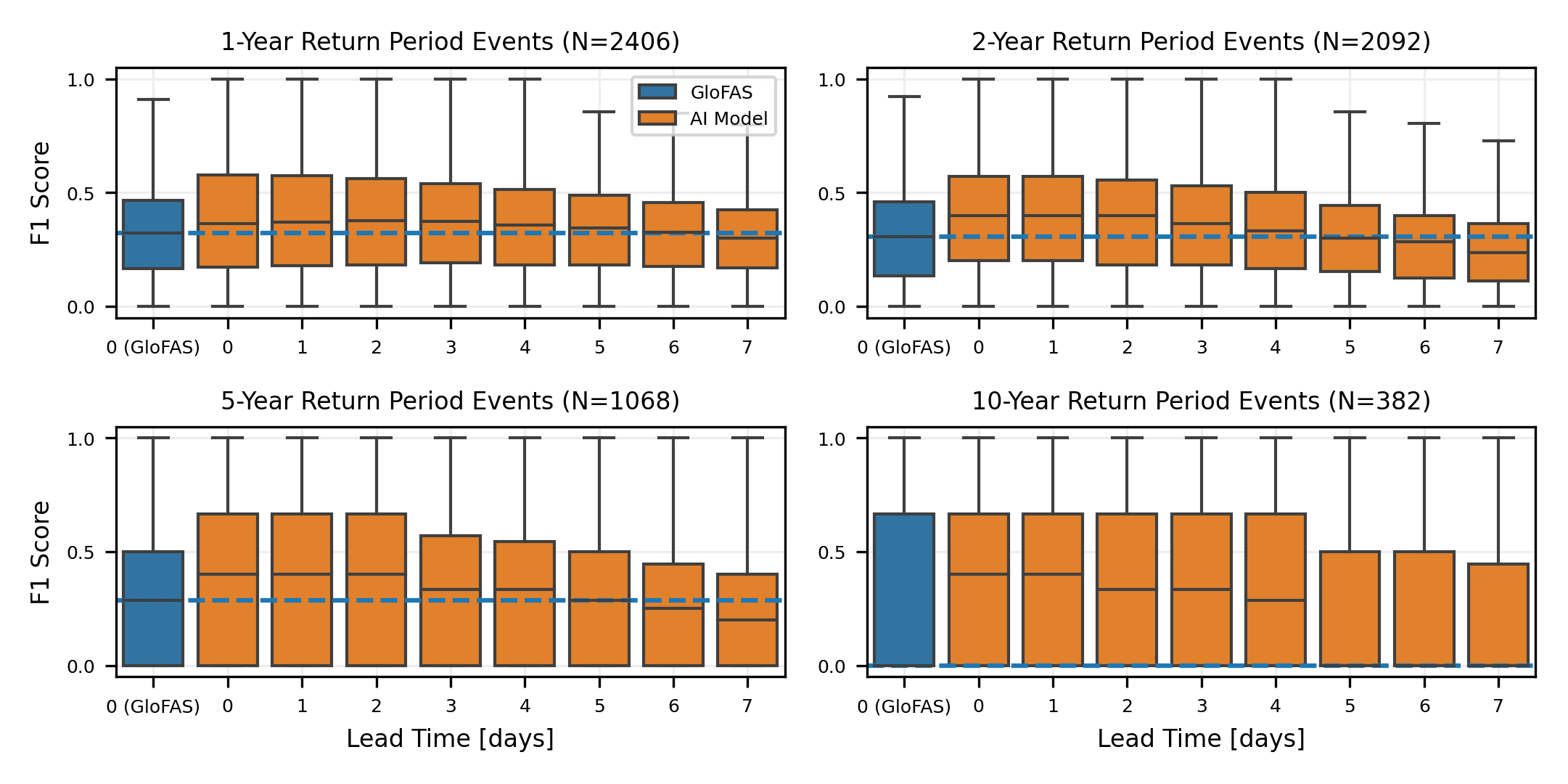}
    \caption{Distributions over F1 scores at all evaluation gauges as a function of lead time for different return periods. The AI model had similar (not statistically different) or better reliability over 1, 2, and 5 year return periods at 5-day lead time than GloFAS at 0-day lead time. Statistical tests are reported in the main text. Boxes show distribution quartiles and whiskers show the full range excluding outliers. The blue dashed line is the median score for GloFAS nowcasts, and is plotted as a reference.}
    \label{fig:lead-time-reliability-distributions}
\end{figure}

Figure \ref{fig:lead-time-reliability-distributions} shows F1 scores over lead times through the 7-day forecast horizon for return periods between 1 and 10 years. Compared with GloFAS nowcasts (0-day lead time), AI forecasts have either better or not statistically different reliability (F1 scores) up to a 5-day lead time for 1-year (AI is significantly better; $N=2,415$, $p=6e-6$, $d=0.08$), 2-year (no statistical difference; $N=2,162$, $p=0.98$, $d=2e-4$), and 5-year (no statistical difference; $N=1,298$, $p=0.69$, $d=0.025$) return period events.

\subsection{Continents} \label{sec:results:continents}
Both models show differences in reliability in different areas of the world. Over 5-year return period events, GloFAS has a 54\% difference between mean F1 scores in the lowest scoring continent (South America: $f1=0.15$) and the highest scoring continent (Europe: $f1=0.32$), meaning that, on average, true positive predictions are twice as likely (at a proportional rate). The AI model also has a 54\% difference between mean F1 scores in the lowest scoring continent (South America: $f1=0.21$) and the highest scoring continent (South West Pacific: $f1=0.46$), which is due mostly to a large increase in skill in the South West Pacific relative to GloFAS ($d=0.68$). 

\begin{figure}[ht]
    \centering
    \includegraphics[width=120mm]{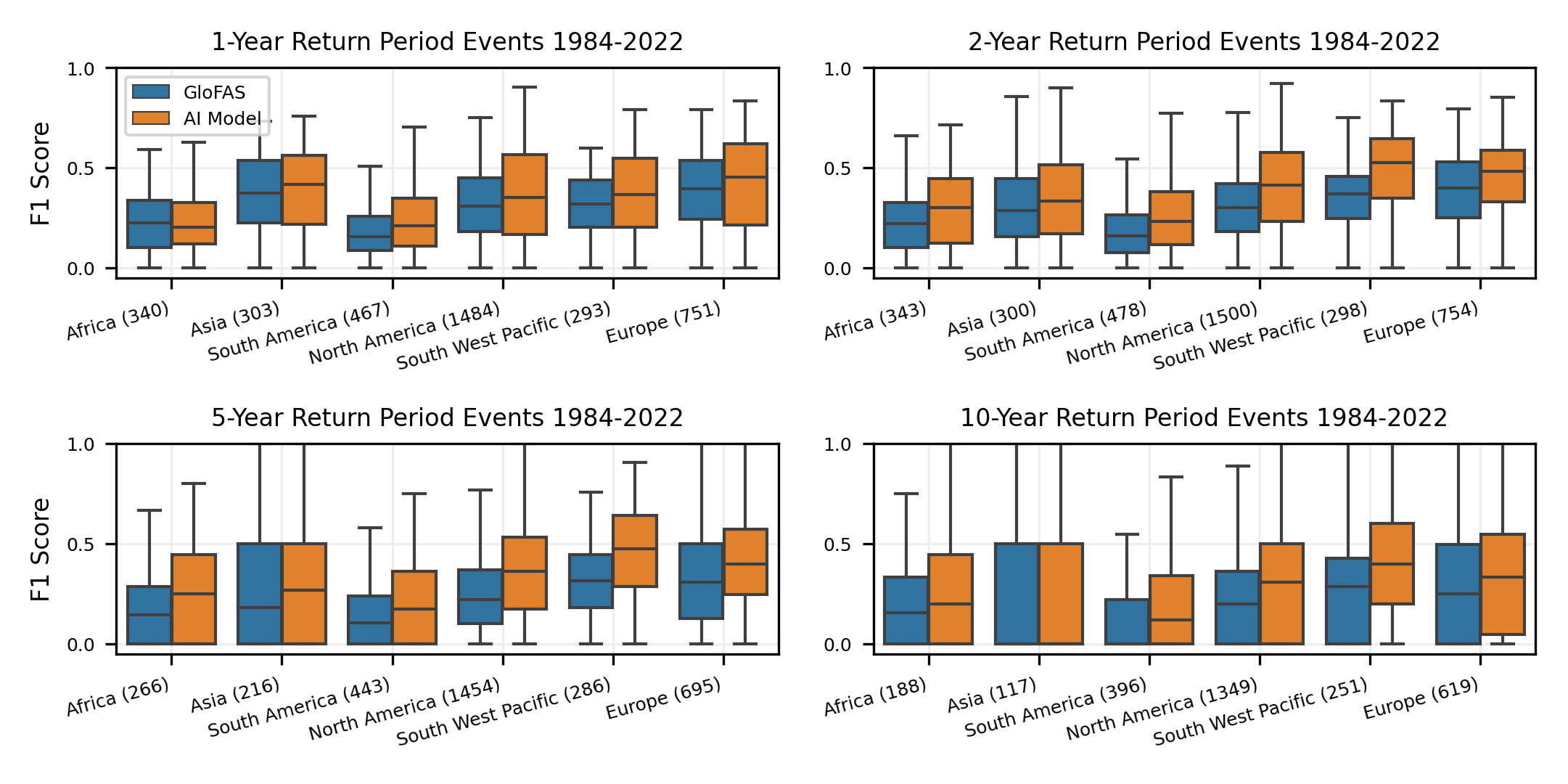}
    \caption{F1 score distributions over different continents and lead times. The AI model has higher scores in all continents and return periods with three exceptions where there is no statistical difference: Africa over 1-year return period events and Asia over 5-year and 10-year return period events. Both models have large location-based differences between reliability that could be addressed by increasing global access to open hydrologic data. Statistical tests are reported in the main text. Boxes show distribution quartiles and whiskers show the full range except outliers.}
    \label{fig:continent-reliability-scores-distributions}
\end{figure}

Figure \ref{fig:continent-reliability-scores-distributions} shows distributions of F1 scores over continents and return periods. The AI model has higher scores in all continents and return periods ($p<1e-2$, $0.10<d<0.68$) with three exceptions where there is no statistical difference: Africa over 1-year return period events ($p=0.07$, $d=0.03$) and Asia over 5-year ($p=0.04$, $d=0.12$) and 10-year ($p=0.18$, $d=0.12$) return period events. 

\section{Is forecast reliability predictable?} 
\label{sec:predictability}
A challenge to forecasting in ungauged basins is that there is often no way to evaluate reliability in locations without ground-truth data. A desirable quality of a model is that forecast skill should be predictable from other observable variables, like mapped or remotely sensed geographical and/or geophysical data. Additionally, while AI-based forecasting offers better reliability in most places, this is not the case everywhere. It would be beneficial to be able to predict where different models can be expected to be more or less reliable.

\begin{figure}[h]
    \centering
    \includegraphics[width=89mm]{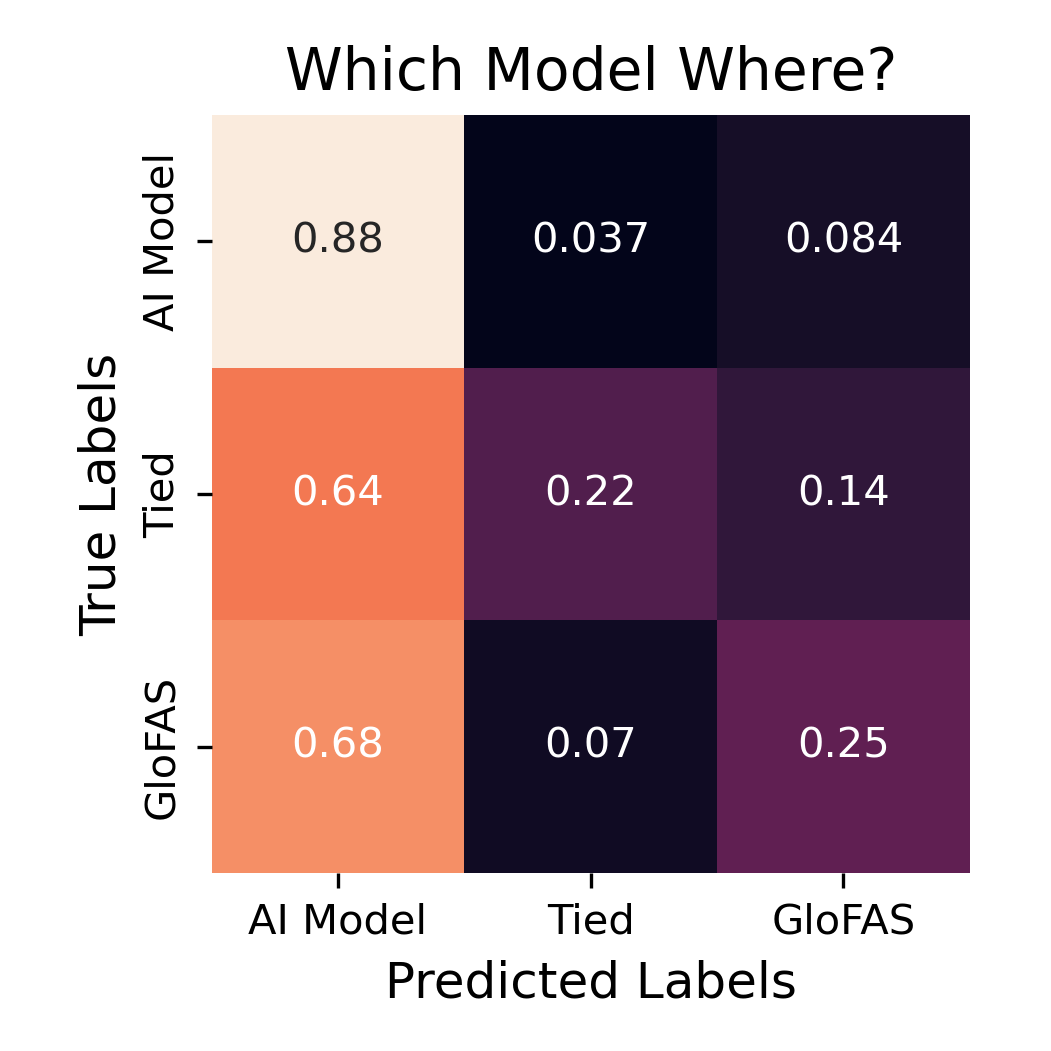}
    \caption{Confusion matrix of a classifier that predicts whether the AI model or GloFAS had a higher (or similar) F1 score in a given watershed based on geophysical catchment attributes ($N=3,360$). We found that this task is generally not possible given available catchment attribute data.}
    \label{fig:which-model-where}
\end{figure}

We have found that it is difficult to use catchment attributes (geographical, geophysical data) to predict where one model performs better than another. Figure \ref{fig:which-model-where} in Extended Data shows a confusion matrix from a random forest classifier trained on a subset of HydroATLAS attributes \cite{linke2019global} that predicts whether the AI model or GloFAS performs better (or similar) in each individual watershed. The classifier was trained with stratified k-fold cross validation and balanced sampling, and usually predicts that the AI model is better (including in 68\% of cases where GloFAS is actually better). This indicates that it is difficult to find systematic patterns about where each model is preferable, based on available catchment attributes.

\begin{figure}[ht]
    \centering
    \includegraphics[width=120mm]{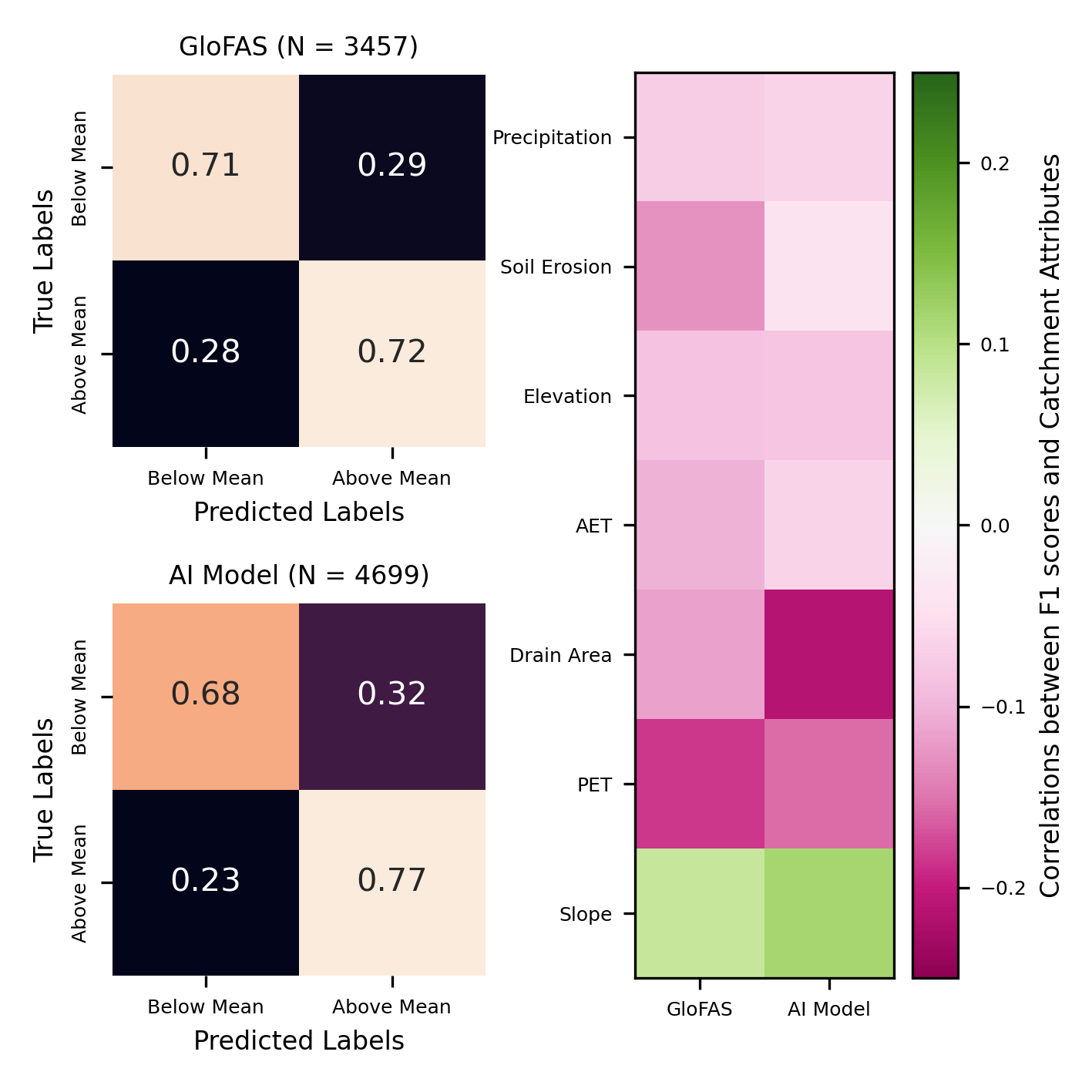}
    \caption{(left) Confusion matrices of out-of-sample predictions about whether F1 scores from GloFAS and the AI model at each gauge are above or below the mean F1 score over all gauges. (right) Correlations between F1 scores and HydroATLAS catchment attributes with the highest feature importance ranks from the trained classifier models.}
    \label{fig:prediction-confusion-matrices}
\end{figure}

On the other hand, it \textit{is} possible to predict, with some skill, where an individual model will perform well vs. poorly. As an example, Figure \ref{fig:prediction-confusion-matrices} shows confusion matrices from random forest classifiers that predict whether F1 scores for out-of-sample gauges (effectively ungauged locations) will be above or below the mean over all evaluation gauges. Both models (the AI model and GloFAS) have similar overall predictability (71\% micro-averaged precision and recall for GloFAS and 73\% for the AI model). 

\begin{figure}[h]
    \centering
    \includegraphics[width=120mm]{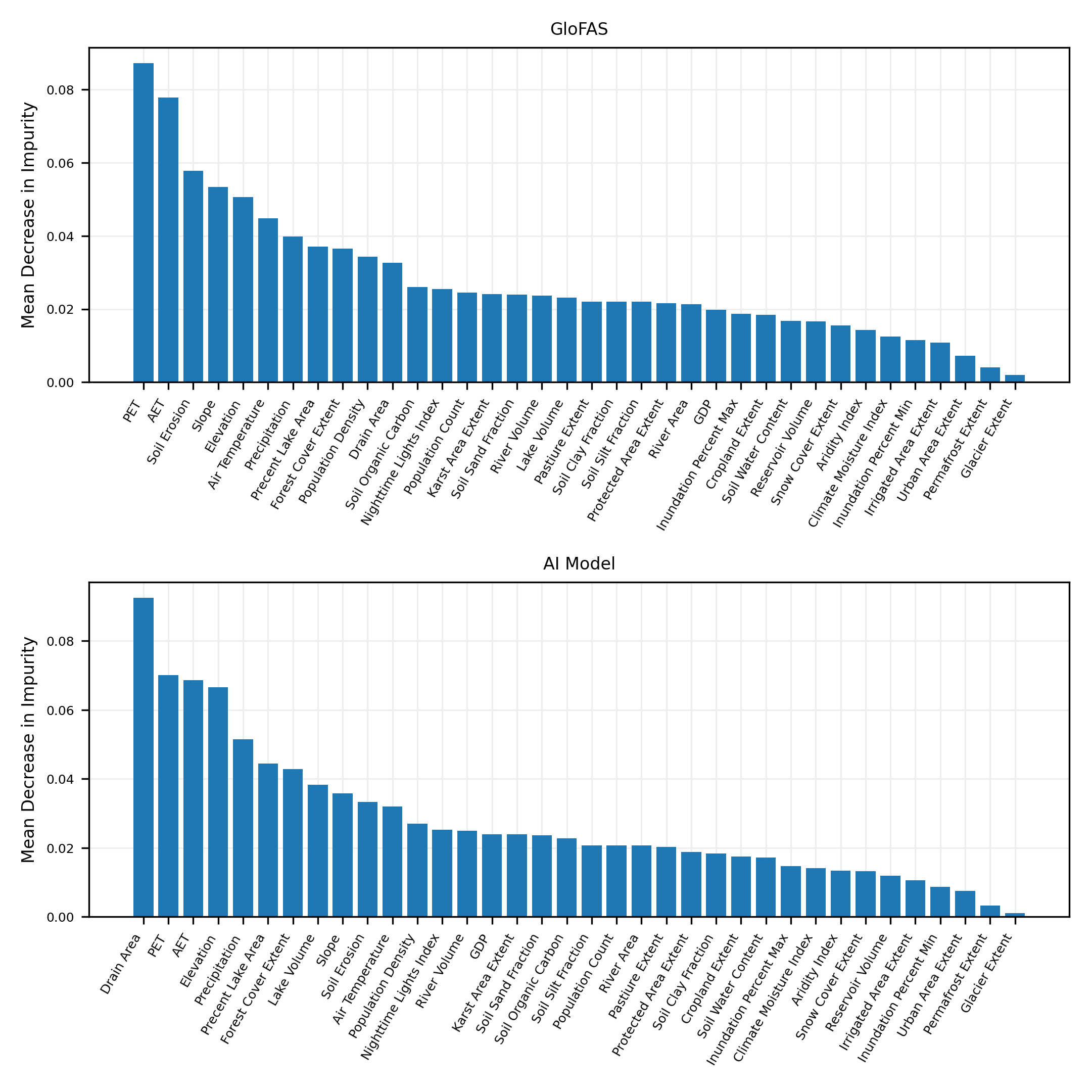}
    \caption{Full feature importance rankings for classifier model in Figure \ref{fig:prediction-confusion-matrices} from Section \ref{sec:predictability} in the main paper.}
    \label{fig:feature-importances}
\end{figure}

Feature importances from these reliability classifiers are shown in Extended Data (Figure \ref{fig:feature-importances}). Feature importance is an indicator about which geophysical attributes determine high vs. low reliability (i.e., what kind of watersheds do these models simulate well vs. poorly). The most important features for the AI model are: drainage area, mean annual potential evapotranspiration (PET), mean annual actual evapotranspiration (AET), and elevation, while the most important features for GloFAS were PET and AET. Correlations between attributes and reliability scores are generally low, indicating a high degree of nonlinearity and/or parameter interaction.

AET and PET are (inverse) indicators of aridity, and hydrology models usually perform better in humid basins because peaky hydrographs that occur in arid watersheds are difficult to simulate. This effect is present for both models. The AI model is more correlated with basin size (drainage area) and generally performs better in smaller basins. This indicates a way that ML-based streamflow modeling might be improved, for example by focusing training or fine tuning on larger basins, or by implementing an explicit routing or graph model to allow for direct modeling of subwatersheds or smaller hydrological response units -- for example as outlined by \cite{kratzert2021large}.

\begin{figure}[ht]
    \centering
    \includegraphics[width=120mm]{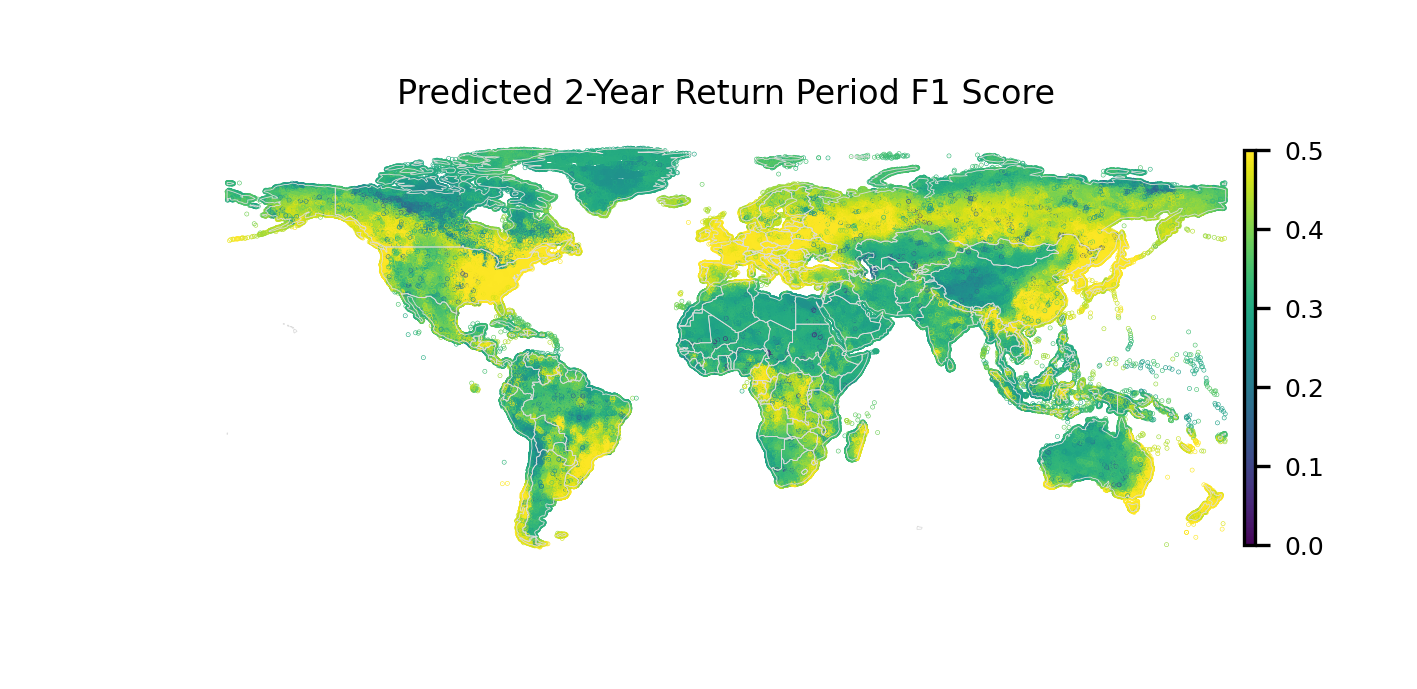}
    \caption{Predictions of 5-year return period F1 scores over 1.03 million HydroBASIN level 12 watersheds for AI forecast model.}
    \label{fig:global_predicted_skill_map}
\end{figure}

A global map of the predicted skill from a regression (rather than classifier) version of this random forest skill predictor is shown in Figure \ref{fig:global_predicted_skill_map} for 1.03 million level 12 HydroBASIN watersheds \cite{lehner2013global}. This gives some indication about where a global version of the ungauged AI forecast model is expected to perform well.

\section{Conclusion and Discussion}\label{sec:discussion}

Although hydrological modeling is a relatively mature area of study, areas of the world that are most vulnerable to flood risks often lack reliable forecasts and early warning systems. Using AI and open data sets, we are able to significantly improve the expected precision, recall, and lead time of short term (0-7 days) forecasts of extreme riverine events. We extended, on average, the reliability of currently-available global nowcasts (lead time 0) to a lead time of 5 days, and we were able to use AI-based forecasting to improve the skill of forecasts in Africa to be similar to what are currently available in Europe.  

Apart from producing accurate forecasts, another aspect of the challenge of providing actionable flood warnings is dissemination of those warnings to individuals and organizations in a timely manner. We support the latter by releasing forecasts publicly in real-time, without cost or barriers to access. We provide open access real-time forecasts to support notifications -- e.g., through CAP (Common Alerting Protocol) and push alerts to personal smartphones, -- and through an open online portal \url{https://g.co/floodhub}. All of the reanalysis and re-forecasts used for this study are included in an open source repository, and a research version of the machine learning model used for this study is available as part of the open source NeuralHydrology repository on GitHub \cite{kratzert2022neuralhydrology}.

There is still a lot of room to improve global flood predictions and early warning systems. Doing so is critical for the wellbeing of millions of people worldwide who's lives (and property) could benefit from timely, actionable flood warnings. We believe that the best way to improve flood forecasts from both data-driven and conceptual modeling approaches is to increase access to data. Hydrologic data is required for training or calibrating accurate hydrology models, and for updating those models in real time \cite[e.g., through data assimilation][]{nearing2022data}. We encourage researchers and organizations with access to streamflow data to contribute to the open source Caravan project \url{https://github.com/kratzert/Caravan} \cite{kratzert2023caravan}.

\newpage
\unnumbered
\section{Methods}
\subsection{AI Model}
\begin{figure}[h]
    \centering
    \includegraphics[width=120mm]{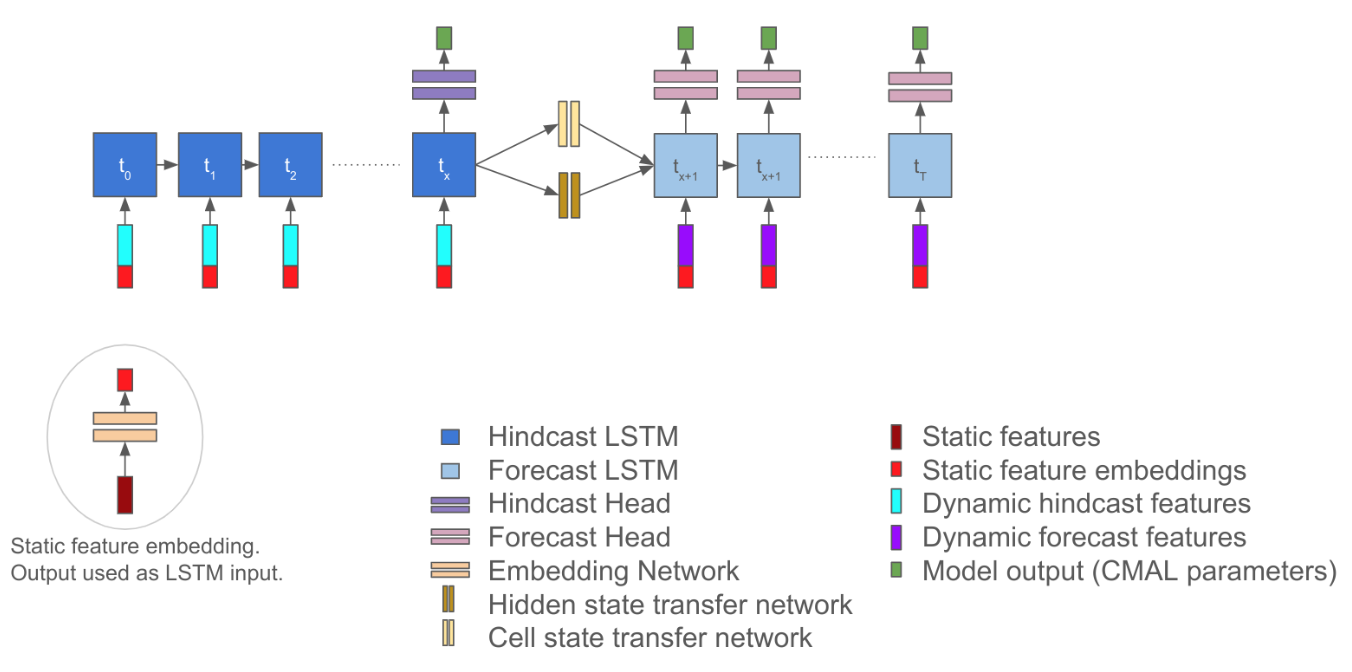}
    \caption{Architecture of the LSTM-based forecast model developed for this project. This is the model used operationally to support the Google Flood Hub \url{https://g.co/floodhub}.}
    \label{fig:model-architecture-diagram}
\end{figure}

The AI streamflow forecasting model reported in this paper extends work by Kratzert et al. \cite{kratzert2019towards}, who developed hydrologic \textit{nowcast} models using Long Short-Term Memory (LSTM) networks that simulate sequences of streamflow data from sequences of meteorological input data. Building on that, developed a \textit{forecast} model that uses an encoder/decoder model with one LSTM running over a historical sequence of meteorological (and geophysical) input data (the encoder LSTM) and another, separate, LSTM that runs over the 7-day forecast horizon with inputs from meteorological forecasts (the decoder LSTM). The model architecture is illustrated in Figure \ref{fig:model-architecture-diagram}.

The model uses a hindcast sequence length of 365 days, meaning that every forecast sequence (0-7 days) saw meteorological input data from the preceding 365 days and meteorological forecast data over the 0-7 day forecast horizon. We used a hidden size of 256 cell states for both the encoder and decoder LSTMs, a linear cell state transfer network and a nonlinear (fully connected layer with hyperbolic tangent activation functions) hidden state transfer network. The model was trained on 50,000 minibatchs with a batch size of 256. All inputs were standardized by subtracting the mean and dividing by the standard deviation of training-period data. 

The model predicts, at each timestep, (timestep-dependent) parameters of a single asymmetric Laplacian distribution over area-normalized streamflow discharge, as described by \cite{klotz2022uncertainty}. The loss function is the joint negative log-likelihood of that heteroscedastic density function. To be clear, the model predicts a separate asymmetric Laplacian distribution at each timestep and each forecast lead time. Results reported in this paper were calculated over a hydrograph that results from averaging the predicted hydrographs from an ensemble of 3 separately-trained encoder/decoder LSTMs. The hydrograph from each of those separately-trained LSTMs is taken as the median (50th percentile) flow value from the predicted Laplacian distribution at each time step and forecast lead time. 

Using the data set described herein, the AI model takes a few hours hours to train on a single NVIDIA-V100 GPU. The exact wall time depends on how often validation is done during training. We use 50 validation steps (every 1,000 batches), resulting in a 10-hour train time for the full global model.

\subsection{Input Data}
The full data set includes model inputs and (streamflow) targets for a total of 152,259 years from 5,680 watersheds. The total size of the data set saved to disk (including missing values in a dense array) is ~60 GB. 

Input data came from the following sources:
\begin{itemize}
    \item Daily-aggregated single-level forecasts from the ECMWF IFS (Integrated Forecast System) HRES (High Resolution) atmospheric model. Variables include: total precipitation (TP), 2-meter temperature (T2M), surface net solar radiation (SSR), surface net thermal radiation (STR), snowfall (SF), and surface pressure (SP).
    \item The same six variables from the ECMWF ERA5-Land reanalysis. 
    \item Precipitation estimates from the NOAA CPC Global Unified Gauge-Based Analysis of Daily Precipitation.
    \item Precipitation estimates from the NASA IMERG (Integrated Multi-satellitE Retrievals for GPM) early run.
    \item Geological, geophysical, and anthropogenic basin attributes from the HydroATLAS database \cite{linke2019global}.
\end{itemize}
All input data were area-weighted averaged over basin polygons over the total upstream area of each gauge or prediction point. The total upstream area for the 5,680 evaluation gauges used in this study ranged from 2.1 to 4,690,998 square kilometers.

No streamflow data were used as inputs to the AI model because (i) realtime data is not available everywhere, especially in ungauged locations, and (ii) because the benchmark (GloFAS) does not use autoregressive inputs. We previously discussed how to use near-realtime target data in an AI-based streamflow model \cite{nearing2022data}. 

\begin{figure}[h]
    \centering
    \includegraphics[width=120mm]{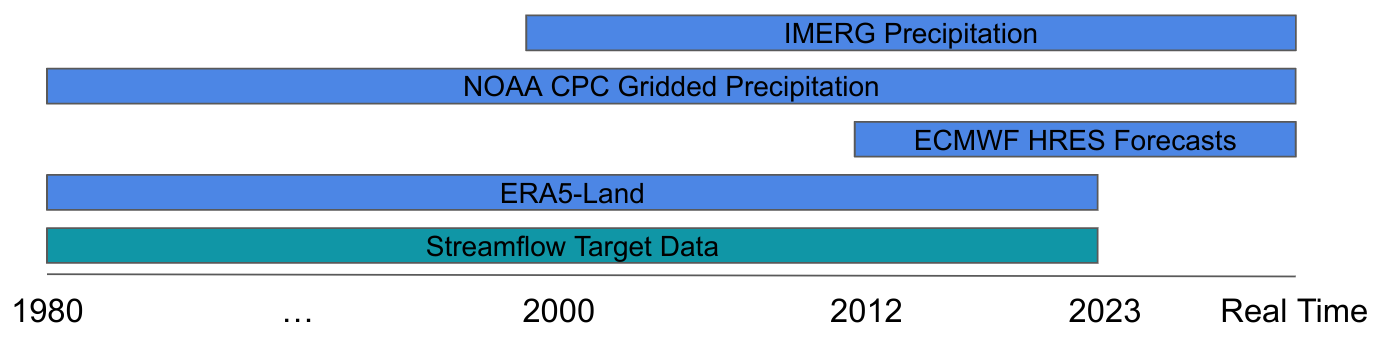}
    \caption{Timeline showing the availability of each data source used for training and prediction with the AI model.}
    \label{fig:data-timeline}
\end{figure}

Figure \ref{fig:data-timeline} shows the time periods of available data from each source. During training, missing data was imputed either by using a similar variable from another data source (e.g., HRES data was imputed with ERA5-Land data), or by imputing with a mean value and then adding a binary flag to indicate an imputed value, as described by Nearing et al. \cite{nearing2022data}.

\subsection{Target \& Evaluation Data}
Training and test targets came from the Global Runoff Data Center (GRDC) \citep{grdc}. Figure \ref{fig:calval_gauge_locations_map} shows the location of all streamflow gauges used in this study for both training and testing. We removed watersheds from the full, public GRDC data set where drainage area reported by GRDC differed by more than 20\% from drainage area calculated using watershed polygons from the HydroBASINS repository -- this was necessary to ensure that poor quality data, due to imperfect catchment delineation, was not used for training. This left us with 5,680 gauges. Since we conducted the experiments reported in this paper, GRDC released catchment polygons for their gauge locations so matching gauges with HydroBASINS watershed boundaries is no longer be necessary.

\subsection{Experiments}
We assessed the performance of the AI model using a set of cross-validation experiments. Data from 5,680 gauges were split in two ways. First, split in time using cross validation folds designed such that no training data from any gauge was used from within 1 year (the sequence length of the LSTM encoder) of any test data from any gauge. Second, split in space using randomized (without replacement) k-fold cross validation with $k=10$. This pair of cross validation processes were repeated so that all data (1984--2021) from all gauges were predicted in a way that was out-of-sample in both time and space. This avoids any potential for data leakage between training to testing. These cross validation experiments are what is reported in the main text of this paper.

Other cross validation experiments are reported briefly in the Extended Data. These include splitting the gauges in time, as above, and in space non-randomly according to the following protocol:

\begin{itemize}
    \item Cross-validation splits across continents (k=6).
    \item Cross-validation splits across climate zones (k=13).
    \item Cross-validation splits across groups of hydrologically separated watersheds (k=8), meaning that no terminal watershed contributed any gauges simultaneously to both training and testing in any cross-validation split.
\end{itemize}

\begin{figure}[h]
    \centering
    \includegraphics[width=120mm]{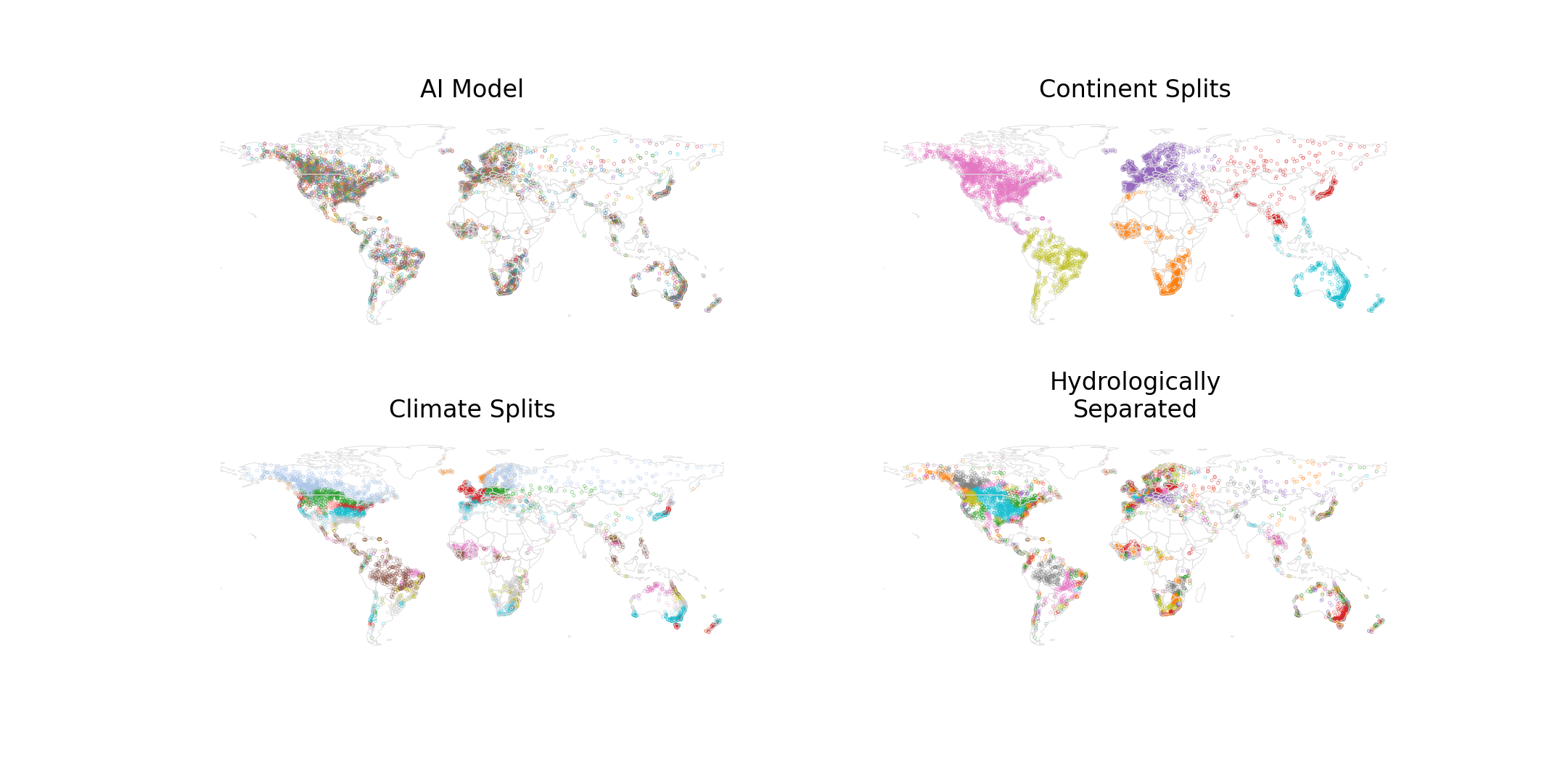}
    \caption{Locations of gauges in each cross-validation split. Different colors in each map represent different cross validation splits. The top-left map shows random splits, which are the results reported in the main text of the paper. The top-right map shows continent splits, so that all basins in a particular continent are in one cross validation group. The bottom-left map shows climate zone splits, so that all basins in each of 13 climate zones are in one cross validation group. The bottom-right map shows splits that group gauges in hydrologically-separated terminal basins.}
    \label{fig:cross_validation_splits_maps}
\end{figure}

The gauges in these cross-validation splits are shown in Figure \ref{fig:cross_validation_splits_maps}. Results from these cross-validation splits are reported in Figures \ref{fig:hydrograph-metrics} and \ref{fig:glofas-calibrated-hydrograph-metrics}.

\subsection{GloFAS}
GloFAS inputs are similar to the input data used in the AI model, with the main differences being:
\begin{itemize}
    \item GloFAS uses ERA5 as forcing data, and not ERA5-Land.
    \item GloFAS (in the data set used here) does not use ECMWF IFS as input to the model. (IFS data are used by the AI model for forecasting only, and we always compare with GloFAS nowcasts).
    \item GloFAS does not use NOAA CPC or NASA IMERG data as direct inputs to the model.
\end{itemize}

GloFAS provides its predictions on a 3 arcminute grid (approx. 5km horizontal resolution). To avoid large discrepancies between the drainage area provided by GRDC and the GloFAS drainage network all GRDC stations with a drainage area smaller than 500 square km were discarded. The remaining gauges  were geolocated on the GloFAS grid and the difference between the drainage area provided by GRDC and the GloFAS drainage network was checked. If the difference between the drainage area was larger than 10\% even after a manual correction of the station location on the GloFAS grid the station was discarded.  A total of 4,090 GRDC stations were geolocated on the GloFAS grid.

\begin{figure}[h]
    \centering
    \includegraphics[width=120mm]{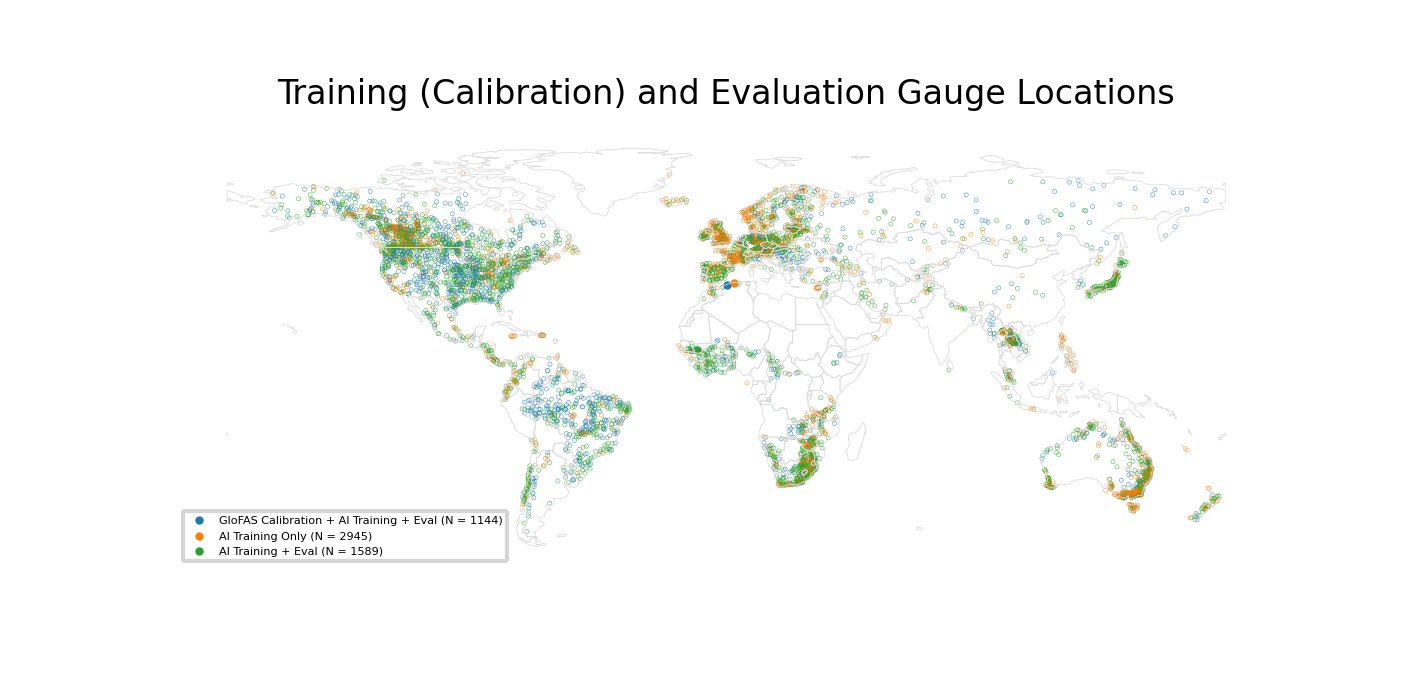}
    \caption{Location of gauges used for (i) training the AI model ($N = 5,860$), (ii) calibrating GloFAS ($N = 1,144$), and (iii) calculating the metrics reported in this paper (evaluation; $N = 4,089$). The AI model is a single model trained on data from all gauges simultaneously, while GloFAS was calibrated separately per-location and following a top-down approach from head-catchments to downstream catchments. All AI model evaluation was done out-of-sample in both location and time. Some of the 5,860 training gauges were excluded from evaluation because it was not possible to match those gauges to a GloFAS pixel.}
    \label{fig:calval_gauge_locations_map}
\end{figure}

Additionally, unlike the AI model, GloFAS was not tested completely out-of-sample. GloFAS predictions came from a combination of gauged and ungauged catchments, and a combination of calibration and validation time periods. Figure \ref{fig:calval_gauge_locations_map} shows the locations of gauges where GloFAS was calibrated. This is necessary because of the computational expense associated with calibrating GloFAS, e.g., over cross-validation splits. More information about GloFAS calibration can be found on the GloFAS Wiki \cite{glofasv4calibrationwiki}.

This means that the comparison with the AI model favors GloFAS. Figure \ref{fig:glofas-calibrated-hydrograph-metrics} shows scores using a set of standard hydrograph metrics in locations where GloFAS is calibrated, and can be compared with Figure \ref{fig:hydrograph-metrics}, which shows the same metrics in all evaluation locations. 

While CEMS releases a full historical reanalysis (without lead times) for GloFAS version 4, long-term archive of reforecasts (forecasts of the past) of GloFAS version 4 do not span the full year at the time of the analysis. Given that reliability metrics must consider the timing of event peaks, this means that it is only possible to benchmark GloFAS at a 0-day lead time.

\subsection{Metrics}
Results in the main text report precision and recall metrics calculated over predictions of events with magnitudes defined by return periods. Precision and recall metrics were calculated separately per gauge for both models. Return periods were calculated separately for each of the 5,680 gauges on both modeled and observed time series (return periods were calculated for observed time series and for modeled time series separately) using the methodology described by the USGS Bulletin 17b \cite{subcommittee1986interagency}. We considered a model to have correctly predicted an event with a given return period if the modeled hydrograph and the observed hydrograph both crossed their respective return period threshold flow values within two days of each other. Precision, recall, and F1 scores were calculated in the standard way separately for each gauge. We emphasize that all models were compared against actual streamflow observations, and it is \text{not} the case that, for example, metrics were calculated directly by comparing hydrographs from the AI model with hydrographs from GloFAS. Note that it is possible for either precision or recall to be undefined for a given model at a given gauge due to there being either no predicted or no observed events of a given magnitude (return period), and it is not always the case that precision is undefined when recall is undefined, and vice-versa. This causes, for example, differences in the precision and recall sample sizes shown in Figure \ref{fig:global-reliability-with-return-period}.

All statistical significance values reported in this paper were assessed using two-sided Wilcoxon (paired) signed-rank tests. Effect sizes are reported as Cohen's term d \citep{sullivan2021using}, which is reported using the convention that the AI model having better mean predictions results in a positive effect size, and vice versa. All box plots show distribution quartiles (i.e., the center bar shows medians, not means) with error bars that span the full range of data excluding outliers. Not all results reported in this paper use all 5,680 gauges due to the fact that some gauges do not have enough samples to calculate precision and recall scores over certain return period events. Sample size is noted for each result.

\begin{figure}[h]
    \centering
    \includegraphics[width=120mm]{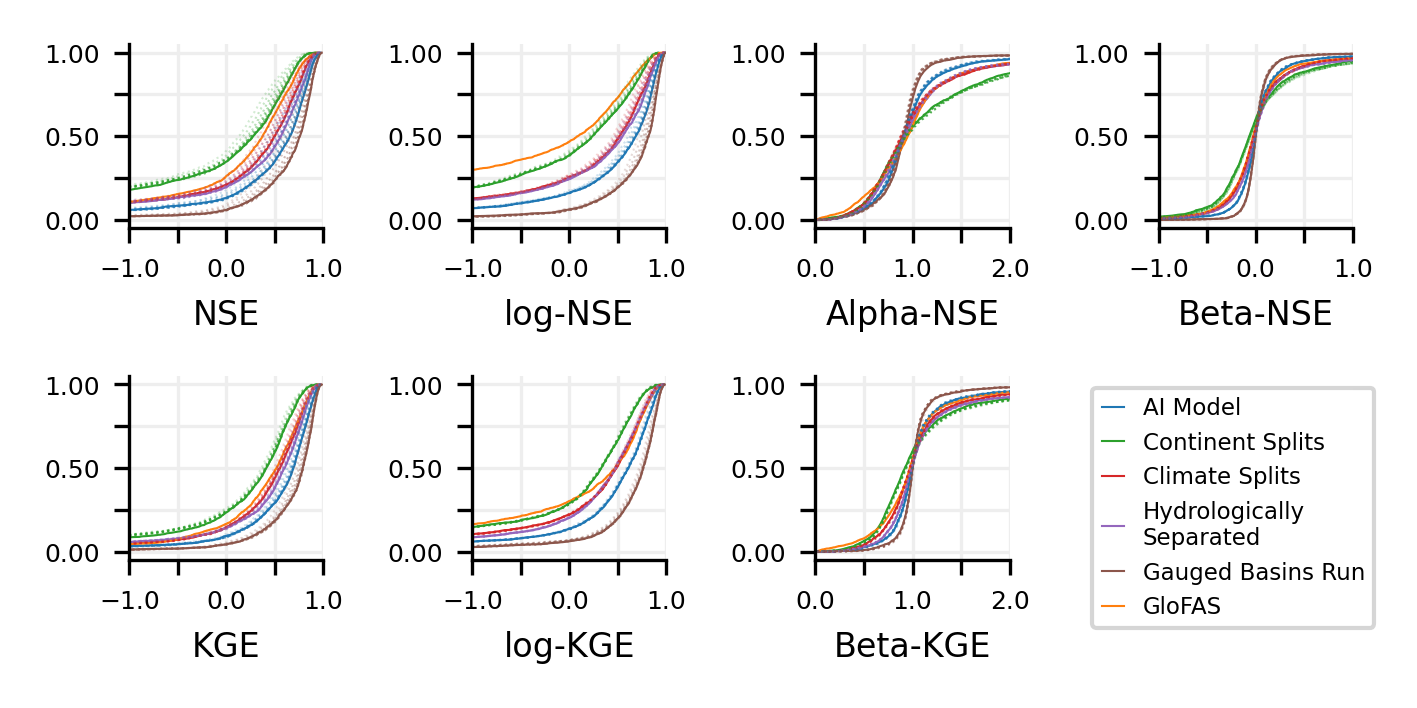}
    \caption{Hydrograph metrics for the AI model and GloFAS over all 4,089 evaluation gauges. Cross validation splits are indicated by colors, and 0 to 7 day lead times are indicated by dashed lines (scores decrease with increasing lead time). Metrics are calculated on the time period 2014-2021. GloFAS shows performance (evaluated using certain metrics) that is similar to the AI model on cross-validation splits that leave entire climate zones are left out of training.}
    \label{fig:hydrograph-metrics}
\end{figure}

\begin{figure}[h]
    \centering
    \includegraphics[width=120mm]{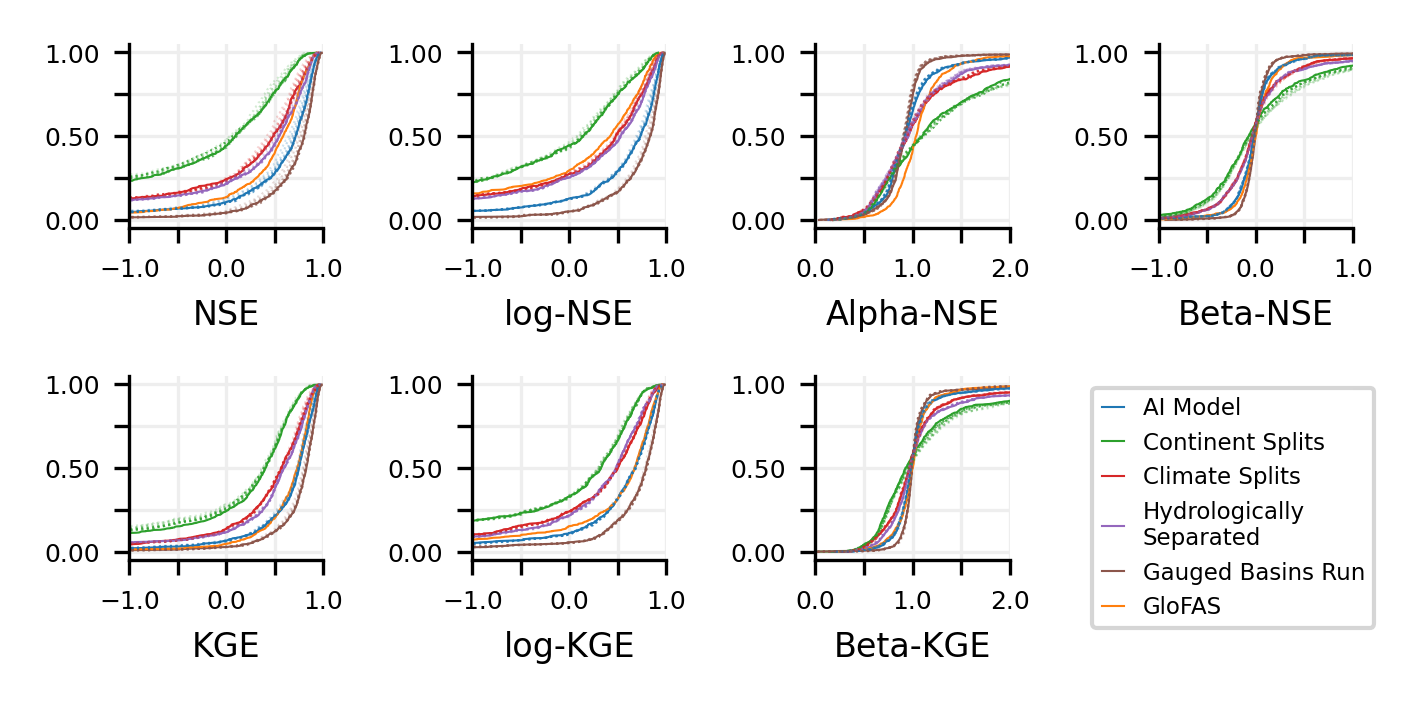}
    \caption{Same as Figure \ref{fig:hydrograph-metrics}, but only for the 1,144 gauges where GloFAS is calibrated. GloFAS is calibrated using the Kling-Gupta Efficiency (KGE), and when evaluated using this metric (as well as bias metrics), shows performance in \textit{gauged} basins that is similar to the AI model in \textit{ungauged} basins.}
    \label{fig:glofas-calibrated-hydrograph-metrics}
\end{figure}

There are a large number of metrics that hydrologists use to assess hydrograph simulations \cite{https://doi.org/10.1029/2022WR033918}, and extreme events in particular \cite{wwrp2016forecast}. Several of these standard metrics are described in Table \ref{tab:hydrograph-metrics} and are reported for the models described in this paper in Figure \ref{fig:hydrograph-metrics}, including bias, Nash--Sutcliffe Efficiency (NSE) \citep{nash1970river}, and Kling--Gupta Efficiency (KGE) \citep{gupta2009decomposition}. KGE is the metric that GloFAS is calibrated to.  Figure \ref{fig:glofas-calibrated-hydrograph-metrics} shows the same metrics, but only calculated over gauges where GloFAS was calibrated (the AI model is still out-of-sample in these gauges). The results in Figures \ref{fig:hydrograph-metrics} and \ref{fig:glofas-calibrated-hydrograph-metrics} show that the ungauged AI model is about as good in \textit{ungauged} basins as GloFAS is in \textit{gauged} basins when evaluated against the metrics that GloFAS is calibrated on (KGE), and is better in ungauged basins than GloFAS is in gauged basins on the (closely related) NSE metrics. However, GloFAS has better overall variance (the Alpha-NSE metric) than the ungauged AI model in locations where it is calibrated (although not in uncalibrated locations), indicating a potential way that the AI model might be improved.

\section*{Data Availability}
Reanalysis (1984--2021) and reforecast (2014--2021) data produced by the AI model for this study, as well as corresponding GloFAS benchmark data are available at \url{https://doi.org/10.5281/zenodo.10045596}. 

Daily river discharge simulations are available for both GloFAS version 3 and GloFAS version 4 from the Climate Data Store \cite{grimaldi2023cds}. For a summary of GloFAS versioning, please see (\url{https://confluence.ecmwf.int/display/CEMS/GloFAS+versioning+system}).

\section*{Code Availability}
The forecasting model developed for this project was integrated into the NeuralHydrology code base \cite{kratzert2022neuralhydrology} that is available at \url{https://neuralhydrology.github.io}. This research code base differs from the operational model that was used in this article primarily in that it can be run on standard compute systems in Linux, iOS, and Windows environments. Training the models reported in this paper with the NeuralHydrology codebase is not plug-and-play.

Code for reproducing the figures and analyses reported in this paper is available at \url{https://github.com/google-research-datasets/global_streamflow_model_paper}. This repository calculates metrics for the AI model and GloFAS outputs, as reported in this paper, and requires the Zenodo data set linked in the Data Availability section below.

\bibliography{bibliography}

\newpage
\section*{Author Contributions}
GN conducted experiments and analyses and wrote the first paper draft that was edited by all coauthors. GS, FK, and OG contributed significantly to experimental design, and the design of figures. All authors contributed to development of the AI model. Authors with ECMWF affiliation (SH, FP, CP) additionally helped to ensure proper processing of GloFAS data. YM supervised the research.

\section*{Author Information}
The authors declare no competing interests. Please contact either the first author (Grey Nearing; \href{mailto:nearing@google.com}{nearing@google.com}) or last author (Avinatan Hassidim; \href{mailto:avinatan@google.com}{avinatan@google.com}) for correspondence and requests, including questions regarding reprints and permissions.

\newpage
\section*{Extended Data}

\begin{table}[h]
\caption{\textbf{A selection of standard hydrograph evaluation metrics}}
\label{tab:hydrograph-metrics}
\begin{tabular*}{\textwidth}{@{\extracolsep\fill}rll}
\toprule
Metric & Description & Reference \\ 
\midrule
NSE & Nash--Sutcliffe efficiency & Eq.\ 3 in \cite{nash1970river} \\
log-NSE & Nash--Sutcliffe efficiency in logarithmic space &  $\sim$ \\
Alpha-NSE & Ratio of standard deviations of observed and simulated flow & Eq.\ 4 in \cite{gupta2009decomposition} \\
Beta-NSE & Bias scaled by standard deviation of observations & Eq.\ 4 in \cite{gupta2009decomposition} \\
KGE & Kling--Gupta efficiency & Eq.\ 9 in \cite{gupta2009decomposition}  \\
log-KGE & Kling--Gupta efficiency in logarithmic space & $\sim$ \\
Beta-KGE & Ratio of mean simulated and mean observed flow & Eq.\ 10 in \cite{gupta2009decomposition} \\
\bottomrule
\end{tabular*}
\end{table}

\newpage

\newpage

\newpage

\newpage

\newpage

\newpage

\newpage

\newpage

\end{document}